\documentclass[runningheads]{llncs}
\usepackage[T1]{fontenc}
\usepackage{graphicx}
\usepackage{amsmath}
\usepackage{amsfonts}
\usepackage{amssymb}
\usepackage{booktabs}
\usepackage[ruled,vlined]{algorithm2e}
\usepackage{xcolor}

\begin{document}
\title{Beyond the Single-Best Model:\\ Rashomon Partial Dependence Profile for Trustworthy Explanations in AutoML\thanks{Accepted at 28th International Conference on Discovery Science 2025}}
\titlerunning{Rashomon Partial Dependence Profile for Trustworthy Explanations}
%
\author{Mustafa Cavus\inst{1}\orcidID{0000-0002-6172-5449} \and
Jan N. van Rijn\inst{2}\orcidID{0000-0003-2898-2168} \and
Przemysław~Biecek\inst{3,4}\orcidID{0000-0001-8423-1823}}
%
%
\institute{Department of Statistics, Eskisehir Technical University, Turkiye\\
\email{mustafacavus@eskisehir.edu.tr} 
\and Leiden Institute of Advanced Computer Science, Leiden University, the Netherlands
\and Faculty of Mathematics and Information Science, Warsaw University of Technology, Poland \and Informatics and Mechanics, University of Warsaw, Faculty of Mathematics, Poland}
\maketitle              
\begin{abstract}
Automated machine learning systems efficiently streamline model selection but often focus on a single best-performing model, overlooking explanation uncertainty—an essential concern in hu\-man-cen\-te\-red explainable AI. To address this, we propose a novel framework that incorporates model multiplicity into explanation generation by aggregating partial dependence profiles (PDP) from a set of near-optimal models, known as the Rashomon set. The resulting Rashomon PDP captures interpretive variability and highlights areas of disagreement, providing users with a richer, uncertainty-aware view of feature effects. To evaluate its usefulness, we introduce two quantitative metrics, the \textit{coverage rate} and the \textit{mean width of confidence intervals}, to evaluate the consistency between the standard PDP and the proposed Rashomon PDP. Experiments on 35 regression datasets from the OpenML-CTR23 benchmark suite show that in most of the cases, the Rashomon PDP covers less than $70\%$ of the best model's PDP, underscoring the limitations of single-model explanations. Our findings suggest that Rashomon PDP improves the reliability and trustworthiness of model interpretations by adding additional information that would otherwise be neglected. This is particularly useful in high-stakes domains where transparency and confidence are critical.

\end{abstract}
\newpage
\section{Introduction}

As artificial intelligence (AI) systems become increasingly embedded in high-stakes domains such as healthcare, finance, and legal applications, the need for explanations that are not only technically accurate but also meaningful to human users has become paramount. Human-centered explainable AI addresses this need by focusing on creating explanations that align with users’ values, expectations, and decision-making contexts \cite{Ehsan2020,Maity2024}. Unlike traditional explainable AI (XAI) approaches that emphasize algorithmic transparency, human-centered XAI seeks to make AI systems more socially intelligible and trustworthy. However, the diversity in user backgrounds and needs present a fundamental challenge: what constitutes a good explanation varies widely among stakeholders \cite{Ehsan2022}. This necessitates interpretability approaches that go beyond performance metrics to also consider how explanations affect user confidence, understanding, and trust \cite{Suffian2023}. In this context, automated machine learning (AutoML) systems—while streamlining model selection and optimization—tend to prioritize predictive performance over interpretability, often resulting in a single best-performing model. 
Even for notable systems that ensemble various well-performing models (such as Auto-sklearn~\cite{Feurer2022} or AutoGluon~\cite{Erickson2020}), it remains non-trivial to utilise these additional models in robust explanations. 
This can obscure uncertainty and limit the explanatory richness required by human-centered XAI, particularly in sensitive, decision-critical domains.

A persistent obstacle in achieving these goals is explanation disagreement, where different explanation methods—or even different models—offer conflicting explanations of the same prediction \cite{Roy2022,Barr2023}. Such inconsistencies can undermine user trust and contribute to epistemic uncertainty \cite{Löfström2024}, particularly in complex models where interpretability is already limited \cite{Krishna2022}. For example, when two explanation methods attribute a model’s cancer diagnosis to entirely different features—such as tumor size versus cell irregularity—it becomes unclear which rationale to trust, amplifying epistemic uncertainty. While disagreement is often viewed as a problem, recent studies \cite{gomes2024,Reingold2024,Schwarzschild2023,vascotto2024} suggest that it can also signal uncertainty and promote more deliberative, cautious decision-making. Thus, rather than eliminating disagreement, it may benefit from frameworks that expose and manage it in ways that support human assessment.

A foundational cause of explanation disagreement lies in the \textit{Rashomon effect}—the existence of many equally accurate models that offer diverging interpretations of the same data \cite{breiman2001}. This effect reveals the multiplicity inherent in machine learning (ML): different models may perform similarly well in terms of test set performance according to a given performance measure, yet vary in structure, fairness, and interpretability \cite{rudin2021}. This raises ethical and practical concerns about relying on a single best model, as the chosen explanation could have significant consequences depending on which model is selected \cite{watsondaniels2023,watsondaniels2024}. Moreover, decision stages such as data pre-processing in the modeling process can produce a Rashomon effect \cite{cavus_2024,cavus_2025}. Therefore, embracing model multiplicity through the Rashomon lens can help expose this uncertainty and provide users with a richer understanding of the model’s behavior and limitations \cite{biecek2024,ganesh2024}. 

In this paper, we propose a novel framework that integrates the Rashomon effect into model interpretation by leveraging partial dependence profiles (PDP), which is a model-agnostic XAI tool that describes the effect of a feature on the predicted outcome \cite{friedman2001greedy}. Rather than relying solely on the top-performing model, our approach considers a set of models with near-optimal performance, known as the Rashomon set. By aggregating the PDPs of these models into what we term the Rashomon PDP, we offer a more comprehensive and uncertainty-aware explanation. This perspective not only highlights where model consensus exists but also where interpretive ambiguity emerges, ultimately supporting more informed and cautious decision-making. Experimental results conducted on 35 real-world datasets using the H2O~\cite{ledell2020h2o} AutoML show that when the Rashomon ratio (a measure of how many models perform nearly as well as the best one) is high, the coverage of the best model’s PDP by the Rashomon PDP tends to be low. In such cases, our proposed framework proves particularly effective, as it better captures the diversity of plausible explanations. The main contributions of this paper are as follows:
\begin{enumerate}
     \item We propose the Rashomon PDP, a novel aggregation method that reflects model multiplicity and explanation uncertainty.
     \item We introduce a metric to quantify the coverage of best-model explanations by the Rashomon PDP, providing a diagnostic for interpretive robustness.
     \item Through extensive empirical evaluation on 35 datasets, we demonstrate that our framework reveals hidden interpretive variability, particularly in high Rashomon ratio settings.
\end{enumerate}

The paper is structured as follows. Section 2 details preliminaries and defines the Rashomon PDP. Section 3 covers the experimental setup, datasets, and metrics. Results are presented in Section 4. Section 5 provides highlighted examples that demonstrate how the framework performs in both high-agreement and high-divergence settings. Finally, Section 6 discusses conclusions, limitations, and future work.
\section{Methods}

Let $\mathcal{D} = \{(\mathbf{x}_i, y_i)\}_{i=1}^n$ be a dataset where each input $\mathbf{x}_i \in \mathbb{R}^p$ represents a $p$-dimensional feature vector and $y_i \in \mathbb{R}$ is a continuous response variable. We denote the $j$-th feature by $x_{ij}$ for the $i$-th instance, and refer to the corresponding feature dimension across the dataset as $X_j$. The primary goal is to train a diverse ensemble of regression models and analyze their behavior using partial dependence profiles, focusing on models within a Rashomon set.

\subsection{Model Training and Rashomon Set Formation}

A collection of regression models $\mathcal{M} = \{M_1, M_2, \ldots, M_K\}$ is generated using an AutoML framework. Each model $M_k$ is trained on the same training data and evaluated on a separate test set. The predictive performance of each model is assessed using an appropriate regression metric $\phi(M_k)$, such as mean squared error or mean absolute error. The best-performing model is identified as:

\begin{equation}
M^* = \arg\min_{M_k \in \mathcal{M}} \phi(M_k).
\end{equation}

\noindent To investigate model multiplicity, we define the \textit{Rashomon set} $\mathcal{R}_\varepsilon \subseteq \mathcal{M}$ as the set of models whose performance is within a predefined tolerance $\varepsilon > 0$ of the best model:

\begin{equation}
\mathcal{R_\varepsilon} = \left\{ M_k \in \mathcal{M} \mid \phi(M_k) \leq \phi(M^*) * (1 + \varepsilon) \right\}.
\end{equation}

\noindent For example, reasonable values for $\varepsilon$ could be $0.05$ or $0.1$, depending on the exact problem domain. This set captures models that perform similarly well, allowing us to study functional variability among near-optimal models. Two common metrics used to characterize this set are \textit{Rashomon set size} and \textit{Rashomon ratio} \cite{semenova2022}.

\subsubsection{Rashomon Set Size} 
refers to the number of models identified within the Rashomon set, i.e.,
$\text{RSS}_\varepsilon = |\mathcal{R}_\varepsilon|$ quantifies how many models achieve performance comparable to the best-performing model under the tolerance $\varepsilon$.

\subsubsection{Rashomon Ratio} 
is the ratio of the Rashomon set size relative to the total set size of trained models $\text{RR}_\varepsilon = |\mathcal{R}_\varepsilon| / |\mathcal{M}|$. This ratio provides a normalized measure of model multiplicity, indicating the fraction of similarly performing models among all candidates. A higher Rashomon ratio suggests greater diversity of well-performing models, which can be further amplified in the presence of data noise \cite{semenova2023}, as noisy conditions often allow a wider range of models to achieve similar performance. 

\subsection{Partial Dependence Profile}

For a selected input feature $X_j$, the \textit{partial dependence profile} (PDP) \cite{friedman2001greedy} under a model $M_k \in \mathcal{R}$ quantifies the marginal effect of $X_j$ on the model’s output. Formally, the PDP is a function defined over the domain of $X_j$, denoted $\mathcal{X}_j$, and describes how the model's prediction changes as $X_j$ varies while averaging out the effects of all other features.

The estimated partial dependence function for feature $X_j$ at value $x \in \mathcal{X}_j$ is given by:

\begin{equation}
\hat{f}_j^{(k)}(x) = \frac{1}{n} \sum_{i=1}^{n} \hat{f}^{(k)}(x, \mathbf{x}_{i,-j}),
\end{equation}

\noindent where $\hat{f}^{(k)}$ is the prediction function of model $M_k$, and $\mathbf{x}_{i,-j}$ denotes the $i$-th input vector with its $j$-th feature replaced by the fixed value $x$.

\subsection{Rashomon Partial Dependence Profile}

The \textit{Rashomon partial dependence profile} aggregates PDPs across all models in the Rashomon set by averaging them, producing a unified profile that reflects the central trend among near-optimal models:

\begin{equation}
\bar{f}_j(x) = \frac{1}{|\mathcal{R}_\varepsilon|} \sum_{M_k \in \mathcal{R}_\varepsilon} \hat{f}_j^{(k)}(x).
\end{equation}

To quantify uncertainty arising from model variability, a nonparametric bootstrap is applied over the Rashomon set. Specifically, $B$ bootstrap samples $\mathcal{R}^{(1)}_\varepsilon$, $\mathcal{R}^{(2)}_\varepsilon, \ldots, \mathcal{R}^{(B)}_\varepsilon$ are drawn with replacement, and for each replicate:

\begin{equation}
\bar{f}_j^{(b)}(x) = \frac{1}{|\mathcal{R}^{(b)}_\varepsilon|} \sum_{M_k \in \mathcal{R}^{(b)}_\varepsilon} \hat{f}_j^{(k)}(x).
\end{equation}

While the Rashomon PDP itself aggregates over multiple near-optimal models, the bootstrap further quantifies the variability of this aggregate due to the finite sample of models in the Rashomon set. This allows us to construct confidence intervals that reflect the stability of the aggregated explanation across different plausible subsets of models.\\

\noindent A pointwise $100(1 - \alpha)\%$ confidence interval for $\bar{f}_j(x)$ is then computed using the percentile method:

\begin{equation}
\text{CI}_j(x) = \left[ Q_{\alpha/2}\left(\{\bar{f}_j^{(b)}(x)\}_{b=1}^B\right),\ Q_{1 - \alpha/2}\left(\{\bar{f}_j^{(b)}(x)\}_{b=1}^B\right) \right],
\end{equation}

\noindent where $Q_p(\{\bar{f}j^{(b)}(x)\}_{b=1}^B)$ denotes the empirical $p$-th quantile of the bootstrap distribution of $\bar{f}_j^{(b)}(x)$. To summarize, the computation of the Rashomon partial dependence profile along with the associated uncertainty quantification via bootstrap is outlined step-by-step in Algorithm~\ref{alg:rpdp}.

\begin{algorithm}[ht]
\label{alg:rpdp}
\caption{Computation of Rashomon partial dependence profile}
\KwIn{Dataset $\mathcal{D}=\{(\mathbf{x}_i,y_i)\}_{i=1}^n$, model set $\mathcal{M}=\{M_1,\ldots,M_K\}$, performance metric $\phi$, tolerance $\varepsilon>0$, feature index $j$, bootstrap iterations $B$, confidence level $\alpha$.}
\KwOut{Rashomon Partial Dependence Profile $\bar{f}_j(x)$ and confidence intervals $\mathrm{CI}_j(x)$.}

Compute performance $\phi(M_k)$ for each model $M_k$ on test data\;
Select best model $M^* \gets \arg\min_{M_k \in \mathcal{M}} \phi(M_k)$\;
Form Rashomon set $\mathcal{R}_\varepsilon \gets \{ M_k \in \mathcal{M} \mid \phi(M_k) \leq \phi(M^*) * (1 + \varepsilon) \}$\;
Define grid $\{x_1,\ldots,x_m\}$ for feature $X_j$\;

\ForEach{$M_k \in \mathcal{R}_\varepsilon$}{
  \ForEach{$x \in \{x_1,\ldots,x_m\}$}{
    Compute $\hat{f}_j^{(k)}(x) = \frac{1}{n}\sum_{i=1}^n \hat{f}^{(k)}(x, \mathbf{x}_{i,-j})$\;
  }
}

\ForEach{$x \in \{x_1,\ldots,x_m\}$}{
  Compute $\bar{f}_j(x) = \frac{1}{|\mathcal{R}_\varepsilon|} \sum_{M_k \in \mathcal{R}_\varepsilon} \hat{f}_j^{(k)}(x)$\;
}

\For{$b=1$ \KwTo $B$}{
  Draw bootstrap sample $\mathcal{R}_\varepsilon^{(b)}$ with replacement from $\mathcal{R}_\varepsilon$\;
  \ForEach{$x \in \{x_1,\ldots,x_m\}$}{
    Compute $\bar{f}_j^{(b)}(x) = \frac{1}{|\mathcal{R}_\varepsilon^{(b)}|} \sum_{M_k \in \mathcal{R}_\varepsilon^{(b)}} \hat{f}_j^{(k)}(x)$\;
  }
}

\ForEach{$x \in \{x_1,\ldots,x_m\}$}{
  Compute confidence interval\\
  $\mathrm{CI}_j(x) = \left[ Q_{\alpha/2}(\{\bar{f}_j^{(b)}(x)\}_{b=1}^B),\ Q_{1-\alpha/2}(\{\bar{f}_j^{(b)}(x)\}_{b=1}^B) \right]$\;
}

\Return{$\bar{f}_j(x)$, $\mathrm{CI}_j(x)$}
\end{algorithm}

\section{Experimental Setup}

This section presents a systematic evaluation of our framework through a series of experiments on diverse regression tasks. We aim to investigate how model multiplicity, as captured by the Rashomon set, influences feature effect estimates and how uncertainty can be quantified through partial dependence analysis.
\subsection{Datasets}

To evaluate the practical utility of our proposed framework, we conduct experiments using the OpenML Curated Tabular Regression benchmark suite 2023 (OpenML-CTR23)~\cite{fischer2023openml}. This benchmark includes 35 tabular regression datasets carefully selected to represent a wide range of domains, dataset sizes, and feature complexities. Building on the design principles of the OpenML-CC18 classification benchmark, the CTR23 suite is specifically tailored to support systematic evaluation in regression contexts. Its diversity allows for robust investigation of model performance and interpretability across varied real-world scenarios.
\subsection{Setup}

We perform all experiments using the H2O AutoML framework~\cite{ledell2020h2o}, an automated machine learning tool that supports a wide range of algorithms, including gradient boosting machines, random forests, generalized linear models, and stacked ensembles. The Rashomon set $R_\varepsilon$ is created using models whose performance is close to that of the best model produced by H2O AutoML, with a tolerance of $\varepsilon = 0.05$ — a threshold commonly adopted in prior studies \cite{müller2023,cavus_2024}. For every dataset in the benchmark suite, we use a consistent configuration: the number of models is limited to \texttt{max\_models = 20} and the total training time is capped at \texttt{max\_runtime\_secs = 360}. This setup ensures fair comparison across datasets while maintaining computational efficiency.

\subsection{Metrics}

To quantify the variability and reliability of feature effect estimates derived from the Rashomon set, we employ two evaluation metrics. The \textit{mean width of confidence intervals} captures the overall uncertainty across models, while the \textit{coverage rate} assesses how well the Rashomon profile aligns with the partial dependence of the best-performing model.

\subsubsection{Mean Width of Confidence Intervals}

To summarize the overall uncertainty in the Rashomon partial dependence profile for a feature $X_j$, we define the \textit{mean width of confidence intervals} (MWCI). Given a grid of input values $\{x_1, x_2, \ldots, x_{n_x}\}$ for $X_j$, the MWCI is calculated as:

\begin{equation}
\text{MWCI}_j = \frac{1}{n_x} \sum_{\ell=1}^{n_x} \left[ Q_{1 - \alpha/2}\left(\{\bar{f}_j^{(b)}(x_\ell)\}\right) - Q_{\alpha/2}\left(\{\bar{f}_j^{(b)}(x_\ell)\}\right) \right].
\end{equation}
\subsubsection{Coverage Rate}

To assess how well the Rashomon partial dependence profile encompasses the predictions of the best-performing model, we define the \textit{coverage rate} (CR). Let $f_j(x)$ denote the partial dependence profile of the best model $M^*$. Then the coverage rate is defined as:

\begin{equation}
\text{CR}_j = \frac{1}{n_x} \sum_{\ell=1}^{n_x} \mathbb{I}\left[ f_j(x_\ell) \in \text{CI}_j(x_\ell) \right]
\end{equation}

\noindent where $\mathbb{I}[\cdot]$ is the indicator function, which is 1 if the value $f_j(x_\ell)$ lies within the Rashomon-based confidence interval $\text{CI}_j(x_\ell)$, and 0 otherwise. This provides a quantitative measure of how consistently the Rashomon profile captures the behavior of the best model.
\section{Results}

Table~\ref{tab:res} presents key metrics to evaluate model performance and uncertainty across different datasets. Best-performing model performance (BMP) reflects the root mean squared error (RMSE) of the top model on each dataset. Model set size (MSS) denotes the total number of models trained per dataset, as this might vary under the fixed time limit that we set. Rashomon set size (RSS$_{0.05}$) indicates how many models achieve performance close to the best-performing model while the tolerance parameter $\varepsilon = 0.05$, thus forming the Rashomon set. The Rashomon ratio (RR$_{0.05}$), defined as the proportion of the Rashomon set relative to the total model set, captures the extent of model multiplicity. A higher Rashomon ratio indicates that a larger subset of models perform similarly well, implying greater model diversity within the Rashomon set, which is important for assessing explanation diversity.

The results highlight considerable variation in model multiplicity and uncertainty. For example, datasets such as \texttt{cars}, \texttt{grid\_stability}, and four more datasets have small Rashomon sets, indicating that only a few models match the top model’s performance, resulting in lower model diversity. Conversely, datasets with higher Rashomon ratios, such as \texttt{cps88wages} and \texttt{health\_insurance}, exhibit many models with similar performance, reflecting greater diversity among well-performing models. Additionally, MWCI and CR reveal higher uncertainty in datasets such as \texttt{energy\_efficiency} and \texttt{fifa}, suggesting more variability in model predictions and wider confidence intervals. These insights emphasize the significance of considering model multiplicity and uncertainty when interpreting model behavior within the Rashomon set.

\begin{table}[h!]
    \centering
    \caption{Set sizes and uncertainty metrics for each dataset. \textbf{BMP}: Best-performing model performance in terms of RMSE, \textbf{MSS}: Model set size, \textbf{RSS$_{0.05}$}: Rashomon set size, \textbf{RR$_{0.05}$}: Rashomon ratio, \textbf{MWCI}: mean width of confidence intervals, \textbf{CR}: coverage rate.}
    \label{tab:res}

    \resizebox{0.9\textwidth}{!}{
    \begin{tabular}{lrrrlrl}\toprule
        \textbf{Dataset} & \textbf{BMP} & \textbf{MSS} & \textbf{RSS$_{0.05}$} & \textbf{RR$_{0.05}$} & \textbf{MWCI} & \textbf{CR}    \\\midrule
        \texttt{abalone}                        & 2.1467       & 19   & 13    & 0.6842    & 0.5646     & 0.4214   \\
        \texttt{airfoil\_self\_noise}           & 1.5680       & 19   & 1     & -         & -          & -  \\
        \texttt{auction\_verification}          & 371.0463     & 19   & 1     & -         & -          & -  \\
        \texttt{brazilian\_houses}              & 307.0835     & 16   & 1     & -         & -          & -  \\
        \texttt{california\_housing}            & 45532.4256   & 13   & 6     & 0.4615    & 8782.0985  & 0.2687   \\
        \texttt{cars}                           & 2056.7710    & 22   & 2     & 0.0909    & 538.4280   & 1   \\
        \texttt{concrete\_compressive\_strength}& 4.1028       & 22   & 5     & 0.2272    & 0.6134     & 0.6937   \\
        \texttt{cps88wages}                     & 443.8287     & 14   & 13    & 0.9285    & 38.5566    & 0.6410   \\
        \texttt{cpu\_activity}                  & 2.1419       & 8    & 4     & 0.5       & 2.0780     & 0.5976   \\
        \texttt{diamonds}                       & 528.6411     & 16   & 7     & 0.4375    & 350.7176   & 0.3416   \\
        \texttt{energy\_efficiency}             & 0.7068       & 22   & 4     & 0.1818    & 0.6040     & 0.84   \\
        \texttt{fifa}                           & 8659.1076    & 16   & 9     & 0.5625    & 1093.6380  & 0.7092   \\
        \texttt{forest\_fires}                  & 112.5122     & 22   & 21    & 0.9545    & 5.9025     & 0.1436   \\
        \texttt{fps\_benchmark}                 & 0.8370       & 15   & 2     & 0.1333    & 0.2242     & 0.9626   \\
        \texttt{geographical\_origin\_of\_music}& 3.8452       & 22   & 3     & 0.1363    & 0.3885     & 0.9782   \\
        \texttt{grid\_stability}                & 0.0081       & 19   & 2     & 0.1052    & 0.0003     & 1   \\
        \texttt{health\_insurance}              & 14.6829      & 19   & 17    & 0.8947    & 1.5426     & 0.7837   \\
        \texttt{kin8nm}                         & 0.1085       & 18   & 2     & 0.1111    & 0.0119     & 1   \\
        \texttt{kings\_county}                  & 110139.3275  & 16   & 5     & 0.3125    & 77076.2159 & 0.4087   \\
        \texttt{miami\_housing}                 & 86377.6786   & 18   & 8     & 0.4444    & 10383.3733 & 0.4695   \\
        \texttt{Moneyball}                      & 22.6084      & 22   & 3     & 0.1363    & 6.9054     & 0.9812   \\
        \texttt{naval\_propulsion\_plant}       & 0.0006       & 17   & 1     & -         & -          & -  \\
        \texttt{physiochemical\_protein}        & 3.5789       & 15   & 6     & 0.4       & 1.2140     & 0.2888   \\
        \texttt{pumadyn32nh}                    & 0.0215       & 15   & 8     & 0.5333    & 0.0004     & 0.7437   \\
        \texttt{QSAR\_fish\_toxicity}           & 0.8382       & 22   & 12    & 0.5454    & 0.1391     & 0.5444   \\
        \texttt{red\_wine}                      & 0.6035       & 22   & 9     & 0.4090    & 0.0401     & 0.7136   \\
        \texttt{sarcos}                         & 2.0706       & 6    & 4     & 0.6667    & 0.8788     & 0.4259   \\
        \texttt{socmob}                         & 11.5282      & 22   & 1     & -         & -          & -  \\
        \texttt{solar\_flare}                   & 0.7593       & 22   & 9     & 0.4090    & 0.0487     & 0.3750   \\
        \texttt{space\_ga}                      & 0.1094       & 19   & 7     & 0.3684    & 0.0440     & 0.75   \\
        \texttt{student\_performance\_por}      & 0.9307       & 22   & 2     & 0.0909    & 0.0312     & 1   \\
        \texttt{superconductivity}              & 9.2580       & 6    & 5     & 0.8333    & 0.6905     & 0.4865   \\
        \texttt{video\_transcoding}             & 1.1825       & 3    & 2     & 0.6667    & 0.4857     & 1   \\
        \texttt{wave\_energy}                   & 17589.2242   & 2    & 1     & -         & -          & -  \\
        \texttt{white\_wine}                    & 0.6060       & 19   & 6     & 0.3157    & 0.0506     & 0.4454   \\\bottomrule
    \end{tabular}}
\end{table}

\begin{figure}[h!]
    \centering
    \includegraphics[width=0.95\linewidth]{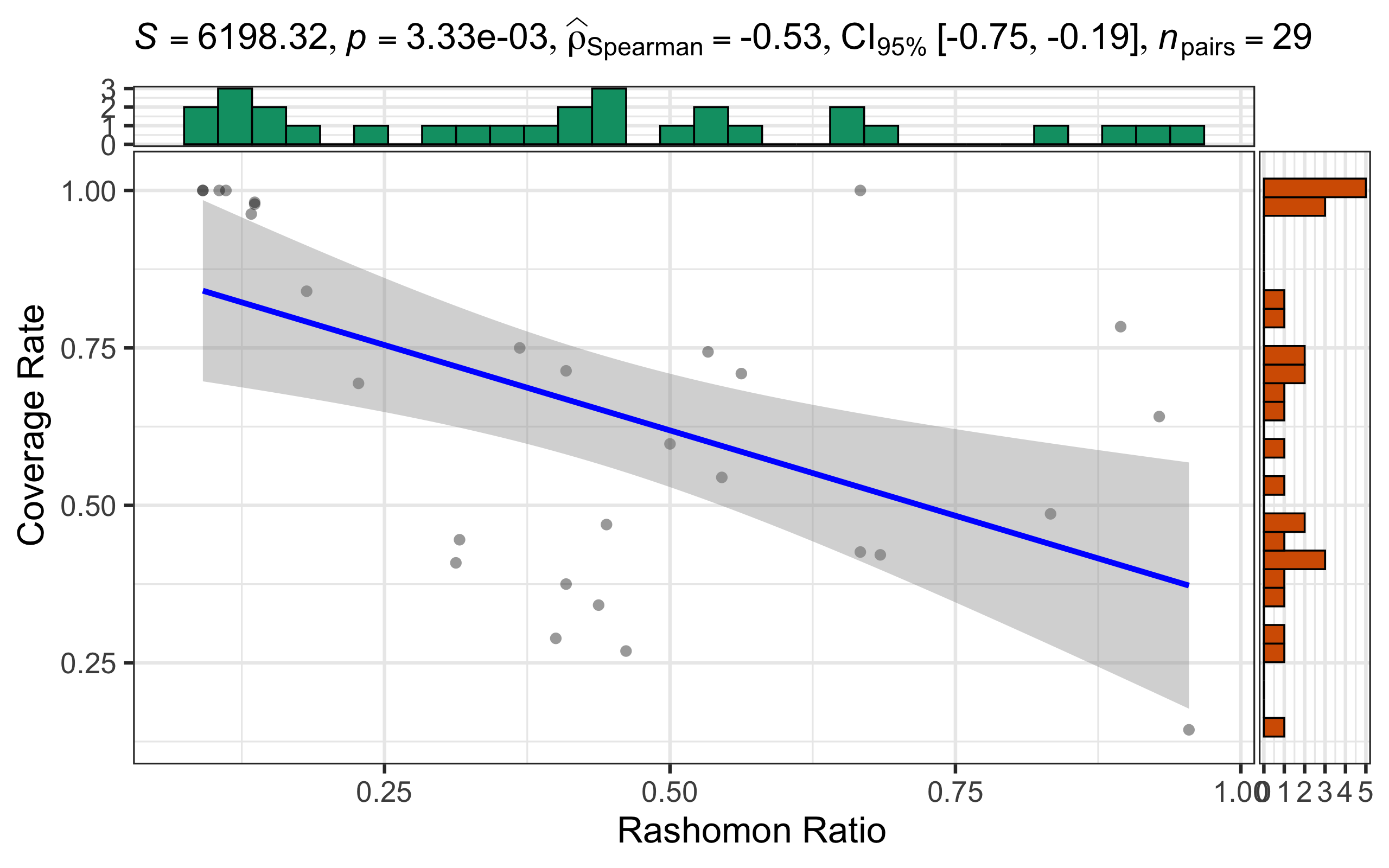}
    \caption{Visual representation of the results across datasets. Each scatter represents a dataset and reveals the Rashomon ratio and coverage rate of that dataset. Most importantly, we have calculated the Spearman’s rank correlation between the Rashomon ratio and coverage rate across datasets.}
    \label{fig:test}
\end{figure}

Figure~\ref{fig:test} illustrates the relationship between the Rashomon ratio and coverage rate, along with Spearman's rank correlation significance test \cite{patil2021ggstatsplot,spearman1987}. We observe a moderate negative correlation, indicating that the coverage rate tends to decrease as the Rashomon ratio increases. This correlation is statistically significant with $\rho = -0.53$, 95\% confidence interval $[-0.75, -0.19]$, and $p = 0.003$. The observed relation between Rashomon ratio and coverage rate aligns with the intuition that as the Rashomon set grows larger, there exists a greater diversity of near-optimal models providing different explanations. This model variability naturally leads to increased epistemic uncertainty, which reduces the reliability of explanations. In other words, when many equally good but diverse models fit the data, pinpointing a consistent explanation becomes more challenging, thereby lowering the coverage rate.

\section{Highlighted Examples}

To illustrate the usability of our framework, we highlight two examples. These examples were selected to illustrate the range of behaviors observed in different Rashomon sets. They highlight how our framework adapts to both high-agreement and high-divergence scenarios, revealing when explanation consensus is trustworthy and when caution is warranted.

\subsection{Example 1: Surface Area on Energy Efficiency Dataset}
\begin{figure}[ht]
    \centering
    \includegraphics[width = 0.95\linewidth]{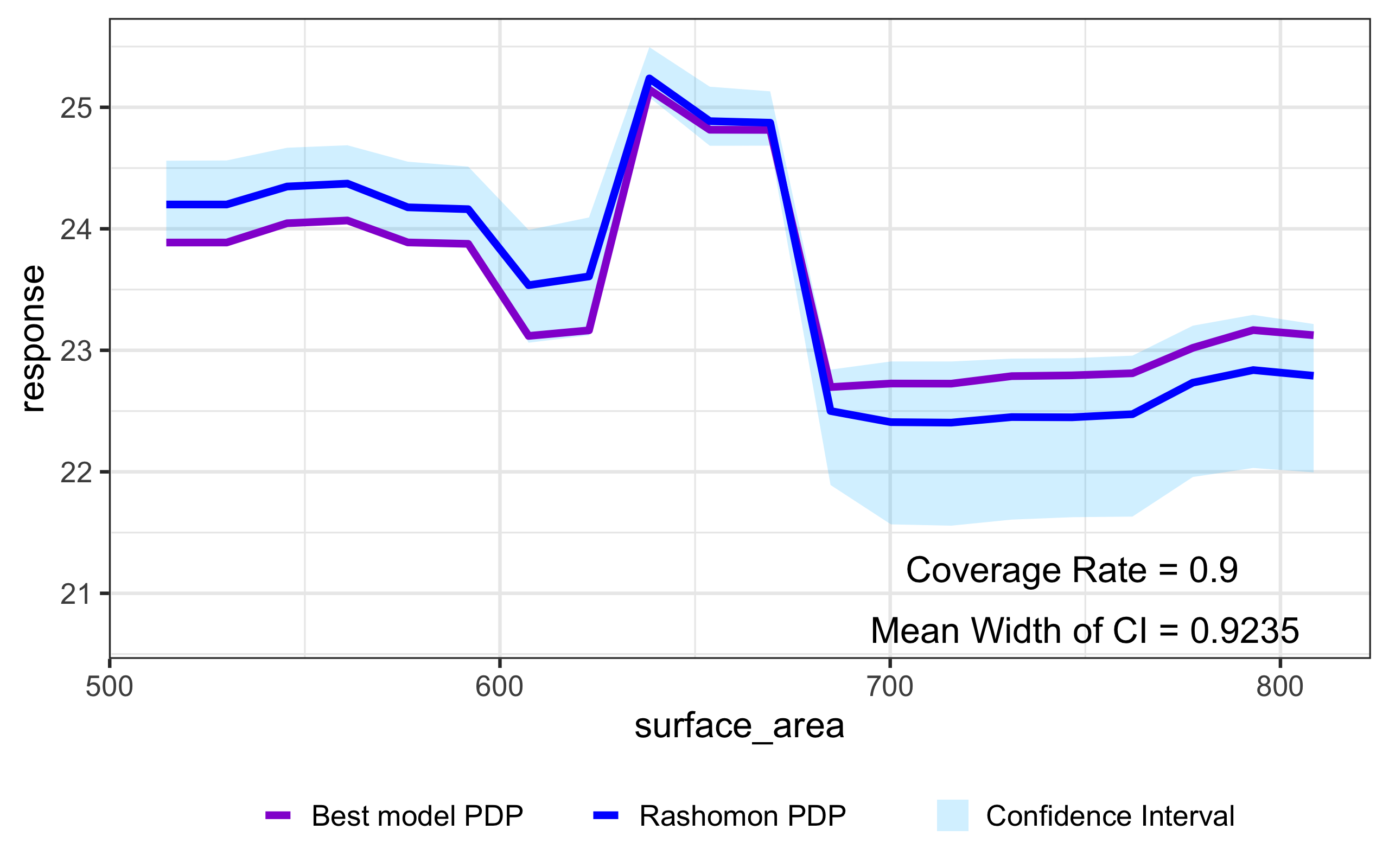}
    \caption{Rashomon PDP on the variable \texttt{surface\_area} of the dataset \texttt{energy\_efficiency}.}
    \label{fig:example1}
\end{figure}

An example Rashomon PDP on the variable \texttt{surface\_area} of the dataset \texttt{e\-ner\-gy\_efficiency} \cite{data1} is presented in Figure~\ref{fig:example1}. 

Predictions remain relatively high and stable for lower \texttt{surface\_area} values, but exhibit a sharp increase around $650$, followed by a steep decline and stabilization at a lower level. The close alignment between the best model PDP and the Rashomon PDP across most of the \texttt{surface\_area} range indicates that the top-performing model behaves similarly to the broader set of well-performing models. Minor deviations, particularly around $600$, suggest localized disagreement without substantial divergence. The confidence interval illustrates the variability within the Rashomon set and remains relatively narrow, with a mean width of $0.9235$, indicating a high level of model agreement. It slightly widens after the sharp drop near $680$, reflecting increased uncertainty in that region. A coverage rate of $0.9$ implies that 90\% of the best model PDP lies within the confidence intervals, further supporting the strong agreement between the best model and the Rashomon PDP. The low mean width of the confidence interval demonstrates that models in the Rashomon set vary little, consistently capturing the relationship between \texttt{surface\_area} and the response variable.

\subsection{Example 2: Temp on Forest Fires Dataset}

\begin{figure}[ht]
    \centering
    \includegraphics[width = 0.95\linewidth]{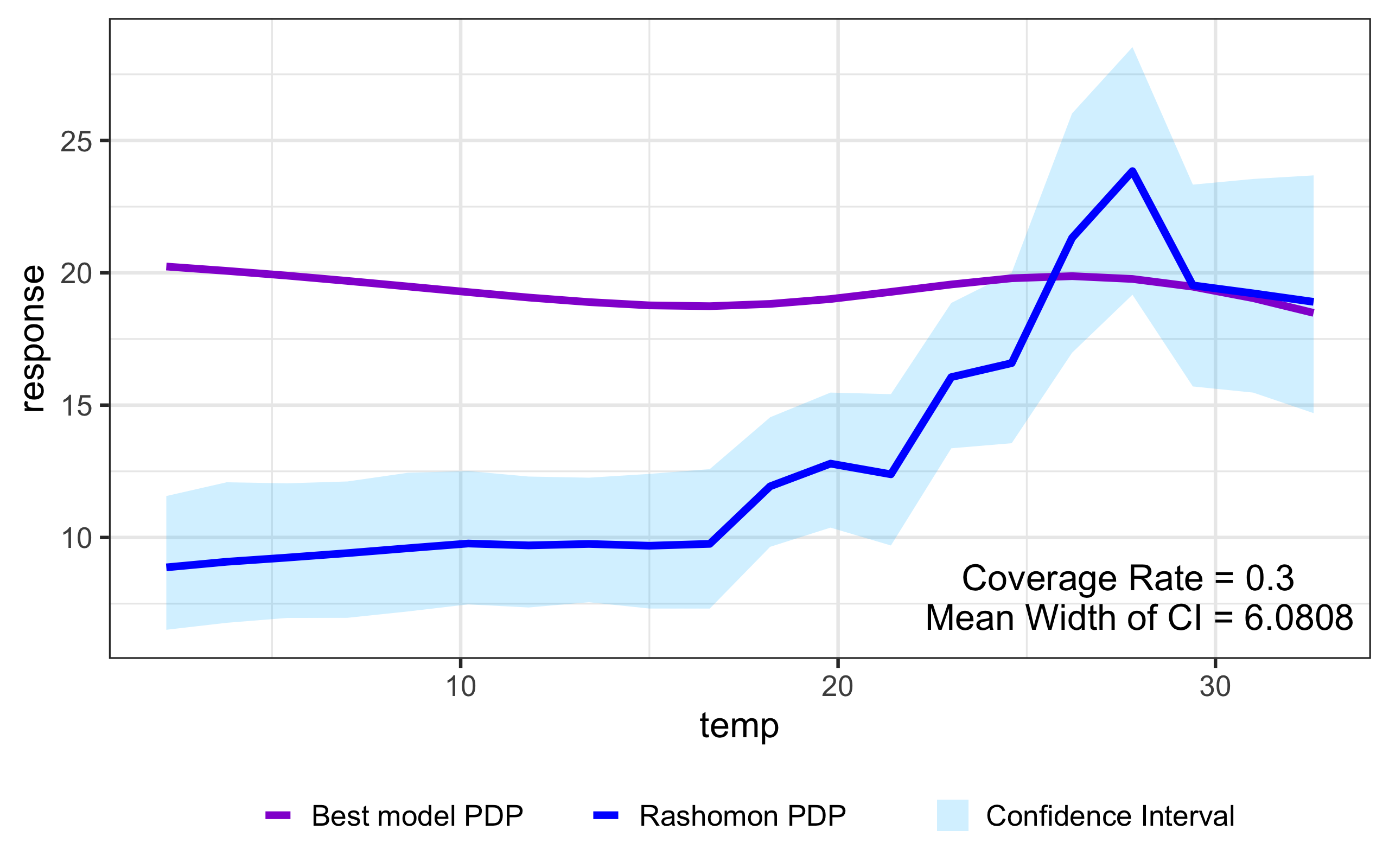}
    \caption{Rashomon PDP on the variable \texttt{temp} of the dataset \texttt{forest\_fires}.}
    \label{fig:example2}
\end{figure}

Another Rashomon PDP example is given on the variable \texttt{temp} of the dataset \texttt{forest\_fires} \cite{data2} given in Figure~\ref{fig:example2}. 

Predictions for \texttt{temp} start low and steadily increase until around 25, followed by greater variability. The best model PDP shows a smooth, slightly decreasing trend, whereas the Rashomon PDP exhibits a pronounced upward trend with higher variance, especially beyond \texttt{temp} values of 20. This divergence indicates that the best model behaves quite differently from many other well-performing models in the Rashomon set, particularly in higher temperature ranges.

The confidence interval around the Rashomon PDP is relatively wide, with a mean width of $6.0808$, reflecting a high level of disagreement among models regarding how \texttt{temp} affects the response. The coverage rate of $0.3$, meaning only $30\%$ of the best model PDP lies within the Rashomon PDP's confidence intervals, further underscores this divergence. This suggests that the best model may be overconfident or misaligned with the broader ensemble in this feature dimension. Such a low coverage rate combined with a high interval width signals significant model uncertainty or instability in the learned relationship between \texttt{temperature} and the response, possibly due to limited or noisy data in this region or inherent complexity in the variable effect. 

Together, these examples demonstrate the practical value of the Rashomon PDP: in cases of strong model agreement, it confirms the reliability of explanations, while in cases of divergence, it surfaces critical uncertainty that would be hidden by relying solely on the best-performing model.

\section{Conclusion}

This paper introduces the Rashomon PDP, a novel approach to explainability in AutoML systems by leveraging the diversity of near-optimal models. Our findings demonstrate that aggregating partial dependence profiles from a Rashomon set offers a reliable understanding of feature effects than relying solely on the best-performing model. Experimental results across diverse regression tasks reveal that explanation variability is both dataset-dependent and influenced by the multiplicity of performant models. Metrics such as mean width of confidence intervals and coverage rate highlight the added value of our method in quantifying uncertainty and enhancing trust in model explanations.

The experimental findings reveal significant variability in the size and diversity of Rashomon sets across different datasets. For instance, datasets such as \texttt{cps88wages} and \texttt{health\_insurance} exhibit high Rashomon ratios, indicating that many models perform similarly well and reflecting greater diversity in model behavior. In contrast, datasets such as \texttt{cars} and \texttt{grid\_stability} have minimal Rashomon sets, suggesting limited model multiplicity and a lower possibility for diverse explanations. Moreover, the examples further illustrate these dynamics: in the \texttt{energy\_efficiency} dataset, the Rashomon PDP shows strong alignment with the best model, indicating consistent behavior across models, whereas in the \texttt{forest\_fires} dataset, a clear divergence is observed, highlighting disagreement among well-performing models. These results underscore the importance of jointly considering model multiplicity and uncertainty when interpreting model behavior and drawing robust conclusions.

Despite these promising results, the study also has several limitations. First, our analysis is restricted to the regression task. 
While the partial dependence profiles are primarily defined for regression tasks, it would be extremely useful to also gain similar explanations for other task types, such as classification or clustering. 
Second, the Rashomon set is formed using a fixed tolerance parameter $\varepsilon$, whose optimal value may vary across datasets and use cases. 
It would be interesting to also explore varying values for epsilon. 
Third, the approach currently focuses on tree-based model ensembles within the H2O AutoML, limiting generalizability to other model families or tools. 
Fourth, the runtime of the AutoML was limited to $360$ seconds per dataset, which may constrain the exploration of the model space and affect the Rashomon set size. Finally, in terms of explainability, the study exclusively employs PDP, potentially overlooking insights that could be gained from alternative XAI methods.

Future work can extend research towards the Rashomon PDP in several directions. Incorporating theoretical background into the Rashomon framework could yield deeper insights into feature contributions. 
Additionally, adaptive or data-driven strategies for Rashomon set construction may improve the robustness of explanations across domains. 
For example, one could investigate the effect of learning curves~\cite{Mohr2024}, and measure whether additional observations sampled in regions of disagreement can improve the coverage rate. 
The approach can also be expanded to include other XAI techniques — such as SHAP, ICE, or LIME — to assess whether the Rashomon perspective reveals similar interpretability patterns across different explanation modalities. 
Finally, user studies are needed to evaluate the impact of Rashomon PDP on human trust, comprehension, and decision-making in real-world applications.

In summary, this work advocates for uncertainty-aware explanations in AutoML, emphasizing that a single-model perspective may be insufficient for trustworthy AI. Rashomon PDP fosters more reliable explanations by accounting for model multiplicity and providing interpretable confidence intervals.

\subsection*{Acknowledgments}
This study is funded by the Polish National Science Centre under the SONATA BIS grant 2019/34/E/ST6/00052.

%
%
%
%

\end{document}